# Bayesian artificial brain with ChatGPT

Renato A. Krohling
UFES - Federal University of Espírito Santo
Labcin - Nature-Inspired Computing Lab &
PPGI  - Graduate Program in Informatics
Av. Fernando Ferrari, 514, Prédio CT X
CEP 29060-270, Vitória, ES
e-mail: krohling.renato@gmail.com

**Abstract:** This paper aims to investigate the mathematical problem-solving capabilities of Chat Generative Pre-Trained Transformer (ChatGPT) in case of Bayesian reasoning. The study draws inspiration from Zhu & Gigerenzer's research in 2006, which posed the question: Can children reason the Bayesian way? In the pursuit of answering this question, a set of 10 Bayesian reasoning problems were presented. The results of their work revealed that children's ability to reason effectively using Bayesian principles is contingent upon a well-structured information representation. In this paper, we present the same set of 10 Bayesian reasoning problems to ChatGPT. Remarkably, the results demonstrate that ChatGPT provides the right solutions to all problems.

## 1. Introduction

The process through which human beings acquire knowledge over the course of their lives has been extensively explored. According to the seminal research conducted by Piaget & Inhelder (1951/1975), it is generally observed that by the age of 12, children possess an understanding of fundamental concepts such as the laws of combinatorics, the law of large numbers, the irreversibility of chance processes, and other essential aspects of probability theory. For the interested reader, see appendix A for a short description of the Bayes Theorem.

Experiments conducted with children from the 4th to the 6th grade in Germany and China (Zhu & Gigerenzer, 2006) indicate that the Bayesian reasoning abilities of children are contingent upon an accurate representation of information. Properly conveying information in a comprehensible manner plays a pivotal role in influencing the way children process and make sense of various concepts.

ChatGPT is a language model developed by OpenAI, built on the GPT (Generative Pre-trained Transformer) architecture. It is designed to generate human-like text based on the input it receives. GPT models are a type of artificial intelligence that have been trained on a massive amount of text data to understand and generate coherent and contextually relevant responses. ChatGPT specifically refers to a version of the GPT model that has been fine-tuned to perform well in generating conversational responses. Although ChatGPT does not explicitly use Bayesian reasoning as part of its core functioning, we show that it can solve Bayesian reasoning problems.

Next, we present the 10 problems of Bayesian reasoning (Zhu & Gigerenzer, 2006) along with the respective solutions provided by ChatGPT.

## 2. Bayesian reasoning problems solving with ChatGPT

### Problem 1
Red nose problem: conditional probabilities
Pingping goes to a small village to ask for directions. In this village, the probability that the person he meets will lie is 10%. If a person lies, the probability that he/she has a red nose is 80%. If a person doesn't lie, the probability that he/she also has a red nose is 10%. Imagine that Pingping meets someone in the village with a red nose. What is the probability that the person will lie?

```
Answer:
   • The probability that a person in the village will lie
     is 10%.
   • If a person lies, the probability that they have a red
     nose is 80%.
   • If a person doesn't lie, the probability that they
     have a red nose is 10%.
   We want to find the probability that a person with a red
nose will lie. This can be calculated using Bayes' theorem:
   P(Lie|Red Nose) = (P(Red Nose|Lie) * P(Lie)) / P(Red
Nose)
   Where:
   • P(Lie|Red Nose) is the probability that a person will
     lie given that they have a red nose.
   • P(Red Nose|Lie) is the probability that a person has a
     red nose given that they lie (80%).
   • P(Lie) is the probability that a person will lie
     (10%).
   • P(Red Nose) is the probability that a person has a red
     nose, which can be calculated using the information
     given:
        • P(Red Nose) = P(Red Nose|Lie) * P(Lie) + P(Red
          Nose|Not Lie) * P(Not Lie)
        • P(Red Nose|Not Lie) is the probability that a
          person has a red nose given that they don't lie
          (10%).
        • P(Not Lie) is the complement of P(Lie), which is
          1 - P(Lie).
   Plugging in the values: P(Red Nose) = (0.8 * 0.1) + (0.1
* 0.9) = 0.08 + 0.09 = 0.17
   Now, we can calculate P(Lie|Red Nose):
   P(Lie|Red Nose) = (0.8 * 0.1) / 0.17 ≈ 0.4706
   So, the probability that a person Pingping meets in the
village with a red nose will lie is approximately 47.06%.
```

Red nose problem: natural frequencies
Pingping goes to a small village to ask for directions. In this village, 10 out of every 100 people will lie. Of the 10 people who lie, 8 have a red nose. Of the remaining 90 people who don't lie, 9 also have a red nose. Imagine that Pingping meets a group of people in the village with red noses. How many of these people will lie?

Answer:
*Out of the people in the village with red noses, the number of them who will lie can be calculated based on the information provided:*
   *From the village's population, 10 out of every 100 people will lie, which is 10% lying rate.*
   *Out of these liars, 8 out of 10 have a red nose.*
   *Additionally, out of the remaining 90 people who don't lie, 9 have a red nose.*
   *This means there are a total of 8 + 9 = 17 people with red noses.*
   *Out of these 17 people with red noses, 8 of them will be liars.*
   *So, out of the group of people with red noses, 8 of them will lie. That means 8/17= 0.47.*

Problems 2 through 10 are given below in terms of natural frequencies

**Problem 2**
There is a large package of sweet or salty cookies with various kinds of shapes. In the package, 20 out of every 100 cookies are salty. Of the 20 salty cookies, 14 are round. Of the remaining 80 sweet cookies, 24 are also round. Imagine you take out a pile of round cookies. How many of them are salty cookies?

Answer:
- *Out of every 100 cookies, 20 are salty.*
- *Among the 20 salty cookies, 14 are round.*
- *There are also 80 sweet cookies in the package.*
- *Among these sweet cookies, 24 are round.*

*We want to find out how many of the round cookies are salty.*
   *The total number of round cookies can be calculated by summing the round salty cookies and the round sweet cookies:*
   *Total round cookies = Round salty cookies + Round sweet cookies Total round cookies = 14 + 24 = 38*
   *The proportion of salty cookies among the round cookies can be calculated based on the proportion of salty cookies among all the cookies:*
   *Proportion of salty cookies among round cookies = (Number of salty cookies / Total number of cookies) * 100 Proportion of salty cookies among round cookies = (20 / 100) * 100 = 20%*
   *Now, we can calculate the number of salty cookies among the round cookies:*
   *Number of salty cookies among round cookies = (Proportion of salty cookies among round cookies / 100) * Total round cookies Number of salty cookies among round cookies = (20 / 100) * 38 = 7.6*

*Since you can't have a fraction of a cookie, we need to round this to the nearest whole number. Therefore, approximately 8 of the round cookies will be salty cookies.*

**Problem 3**
The principal of a school announced and explained a new school rule to all the students gathering together on the playground. Then the principal said: "Those who understand what I mean, please put up your hands." 70 out of every 100 students understood. Of these 70 who understood, 63 put up their hands. Of the remaining 30 who didn't understand, 9 put up their hands. Imagine a group of students who put up their hands. How many of them understood the principal?

Answer:
- *Out of the group of students who put up their hands, the number of students who understood the principal can be calculated based on the information provided:*
- *70 out of every 100 students understood the principal's explanation, which means a 70% understanding rate.*
- *Out of these 70 students who understood, 63 put up their hands.*
- *Additionally, out of the remaining 30 students who didn't understand, 9 put up their hands.*

*So, among the group of students who raised their hands, the number who understood the principal is 63 students.*

**Problem 4**
20 out of every 100 children in a school have bad teeth. Of these 20 children who have bad teeth, 10 love to eat sweet food. Of the remaining 80 children who don't have bad teeth, 24 also like to eat sweet food. Here is a group of children from this school who love to eat sweet food. How many of them may have bad teeth?

Answer:
- 20 out of every 100 children have bad teeth.
- Of the 20 children with bad teeth, 10 love to eat sweet food.
- *There are 80 children who don't have bad teeth.*
- Of the 80 children without bad teeth, 24 also like to eat sweet food.

*Now, you're asking about a group of children who love to eat sweet food and how many of them may have bad teeth. Let's calculate:*
*First, let's find out how many children love to eat sweet food in total:*
- *10 children with bad teeth love sweet food.*
- *24 children without bad teeth love sweet food.*

*Total children who love sweet food = 10 + 24 = 34.*
*Now, out of these 34 children who love sweet food, we need to find out how many may have bad teeth. We know that there are 20 children with bad teeth in total. However,*

some of them also love to eat sweet food. We need to
subtract those from the total:
    Children with bad teeth who love sweet food = 10.
    So, the number of children who love sweet food and may
have bad teeth is 10.
    Therefore, 10 children in the group that loves sweet
food may have bad teeth.

**Problem 5**

To protect their children's eyes, mothers always urge children not to watch too much TV. Suppose you want to test this belief and get the following information: 30 out of every 100 children become near-sighted. Of these 30 near-sighted children, 21 of them watch too much TV. Of those 70 children with normal sight, 28 of them watch too much TV. Suppose you meet a group of children who watch too much TV, how many of them may become near-sighted?

Answer:
  • *Out of every 100 children, 30 become near-sighted. This means a 30% occurrence rate of near-sightedness.*
  • *Among these 30 near-sighted children, 21 of them watch too much TV.*
  • *There are 70 children with normal sight.*
  • *Out of these 70 children with normal sight, 28 of them watch too much TV.*
    Now, we want to find out how many of the children who watch too much TV may become near-sighted.
    First, we need to calculate the proportion of near-sighted children among those who watch too much TV:
    Proportion of near-sighted children among TV watchers = (Number of near-sighted TV watchers / Total number of TV watchers) * 100
    Number of near-sighted TV watchers = Number of near-sighted children who watch TV + Number of normal-sighted children who watch TV
    Number of near-sighted TV watchers = 21 + 28 = 49
    Total number of TV watchers = Number of near-sighted children who watch TV + Number of normal-sighted children who watch TV
    Total number of TV watchers = 49 + 28 = 77
    Proportion of near-sighted children among TV watchers = (49 / 77) * 100 ≈ 63.64%
    So, among the group of children who watch too much TV, approximately 63.64% of them may become near-sighted.

**Problem 6**

In Dongdong's town, 10 out of every 100 children are overweight. Of the 10 overweight children, 3 of them have overweight mothers. Of the remaining 90 children who have normal weight, 18 of them still have overweight mothers. Suppose you meet a group of overweight mothers in the town. How many of them have overweight children?

Answer:
> *20 Out of every 100 children in Dongdong's town, 10 are overweight. This means a 10% occurrence rate of overweight children.*
> - *Among these 10 overweight children, 3 of them have overweight mothers.*
> - *There are 90 children with normal weight.*
> - *Out of these 90 children with normal weight, 18 of them have overweight mothers.*
>
> *Now, we want to find out how many of the overweight mothers have overweight children.*
>
> *First, we need to calculate the proportion of overweight children among the children of overweight mothers:*
>
> *Proportion of overweight children among children of overweight mothers = (Number of overweight children of overweight mothers / Total number of children of overweight mothers) * 100*
>
> *Number of overweight children of overweight mothers = Number of overweight children with overweight mothers + Number of normal weight children with overweight mothers*
>
> *Number of overweight children of overweight mothers = 3 + 18 = 21*
>
> *Total number of children of overweight mothers = Number of overweight children with overweight mothers + Number of normal weight children with overweight mothers*
>
> *Total number of children of overweight mothers = 21 + 18 = 39*
>
> *Proportion of overweight children among children of overweight mothers = (21 / 39) * 100 ≈ 53.85%*
>
> *So, among the group of overweight mothers in the town, approximately 53.85% of them have overweight children.*

**Problem 7**

A group of children are playing games with cards. Those who get a card with a picture of a cat on the inner side win a piece of candy. 30 of every 100 cards have a cat picture on one side. Of the 30 cards with a cat picture, 12 of them are red on the other side. Of the remaining 70 cards that have no cat pictures, 35 of them are still red on the other side. Imagine Dingding takes out a group of red cards. How many of them have a cat picture on the other side?

Answer:
> - *Out of every 100 cards, 30 have a cat picture on one side. This means a 30% occurrence rate of cat picture cards.*
> - *Among these 30 cards with a cat picture, 12 of them are red on the other side.*
> - *There are 70 cards that have no cat pictures.*
> - *Out of these 70 cards, 35 of them are red on the other side.*
>
> *Now, we want to find out how many of the red cards have a cat picture on the other side.*

*First, we need to calculate the proportion of cards with cat pictures among the red cards:*
*Proportion of cards with cat pictures among red cards = (Number of cards with cat pictures and red on the other side / Total number of red cards) \* 100*
*Number of cards with cat pictures and red on the other side = Number of cards with cat pictures and red on the other side + Number of cards without cat pictures and red on the other side*
*Number of cards with cat pictures and red on the other side = 12*
*Total number of red cards = Number of cards with cat pictures and red on the other side + Number of cards without cat pictures and red on the other side*
*Total number of red cards = 12 + 35 = 47*
*Proportion of cards with cat pictures among red cards = (12 / 47) \* 100 ≈ 25.53%*
*So, among the group of red cards that Dingding takes out, approximately 25.53% of them have a cat picture on the other side.*

**Problem 8**
In a cold winter in a town, 40 out of every 100 people hurt their hands by the cold. Of the 40 people who hurt their hands, 36 wear gloves in the open air. Of the remaining 60 people with normal hands, 30 also wear gloves. Suppose you meet a group of people who wear gloves in the town. How many of them hurt their hands?

Answer:
- *Out of every 100 people, 40 hurt their hands due to the cold.*
- *Out of the 40 people who hurt their hands, 36 wear gloves in the open air.*
- *This means that out of the 40 people who hurt their hands, 40 – 36 = 4 do not wear gloves.*
- *Out of the remaining 60 people with normal hands, 30 also wear gloves.*
- *This means that out of the 60 people with normal hands, 60 – 30 = 30 do not wear gloves.*

*Now, you're meeting a group of people who wear gloves. This group includes both those who hurt their hands and those with normal hands.*
*The total number of people wearing gloves is 36 (who hurt their hands) + 30 (with normal hands) = 66.*
*Out of these 66 people wearing gloves, 36 people hurt their hands, and 30 people have normal hands.*
*Therefore, the number of people wearing gloves who also hurt their hands is 36.*
*So, if you meet a group of people wearing gloves, 36 of them have hurt their hands due to the cold.*

**Problem 9**
In a hospital, 60 out of every 100 patients get a cold. Of the 60 patients who get a cold, 42 have a headache. Of the remaining 40 patients with other diseases, 12 also have a headache. Suppose you meet a group of patients who have a headache in a hospital. How many of them get a cold?

```
Answer:
   • Out of every 100 patients, 60 get a cold.
   • Out of the 60 patients with a cold, 42 have a
     headache.
   • This means that out of the 60 patients with a cold, 60
     - 42 = 18 do not have a headache.
   • Out of the remaining 40 patients with other diseases,
     12 also have a headache.
   • This means that out of the 40 patients with other
     diseases, 40 - 12 = 28 do not have a headache.
   Now, you're meeting a group of patients who have a
headache. This group includes patients with a headache from
both those who have a cold and those with other diseases.
   The total number of patients with a headache is 42 (from
those with a cold) + 12 (from other diseases) = 54.
   Out of these 54 patients with a headache, 42 patients
have a headache and a cold, and 12 have a headache and
another disease.
   Therefore, the number of patients with a headache who
also have a cold is 42.
   So, if you meet a group of patients with a headache, 42
of them have a cold.
```

**Problem 10**
On a campus, 90 out of every 100 young people you meet are college students of this university. Of the 90 college students, 45 wear glasses. Of the remaining 10 young people that are not students of the university, 3 also wear glasses. Suppose you meet a group of young people who wear glasses on the campus. How many of them are students at this university?

```
Answer
   • Out of every 100 young people on the campus, 90 are
     college students of this university. This means a 90%
     occurrence rate of college students.
   • Among these 90 college students, 45 of them wear
     glasses.
   • There are 10 young people who are not students of the
     university.
   • Out of these 10 non-university students, 3 of them
     wear glasses.
   Now, we want to find out how many of the young people
who wear glasses are students at this university.
   First, we need to calculate the proportion of university
students among those who wear glasses:
```

```
  Proportion of university students among those who wear
glasses = (Number of university students who wear glasses /
Total number of people who wear glasses) * 100
  Number of university students who wear glasses = Number
of college students who wear glasses
  Number of university students who wear glasses = 45
  Total number of people who wear glasses = Number of
college students who wear glasses + Number of non-
university students who wear glasses
  Total number of people who wear glasses = 45 + 3 = 48
  Proportion of university students among those who wear
glasses = (45 / 48) * 100 ≈ 93.75%
  So, among the group of young people who wear glasses on
the campus, approximately 93.75% of them are students at
this university.
```

### 3. Concluding remarks

In this paper, we show that ChatGPT is able to solve all 10 problems involving Bayesian reasoning presented in the form of natural frequencies and also problem 1 (red nose) presented in the form of conditional probability. For us humans, representing information in the form of frequencies rather than conditional probabilities is much more natural for structuring solutions of Bayesian reasoning problems. Regardless of the information provided to ChatGPT, it represents the information in numbers and, as representation is a part of computation, it manages to make correct inferences for the set of problems presented (Zhu & Gigerenzer, 2006). As pointed out by Mingard (2020) neural networks are fundamentally (almost) Bayesian.

Considering that a significant number of elementary school mathematics teachers lack strong mathematical backgrounds, the usage of ChatGPT could prove beneficial. By providing structured and articulated solutions, as well as representing mathematical concepts, ChatGPT could enhance the learning process beyond solely relying on intuitive explanations. This technology has the potential to bridge gaps in mathematical education and support educators in fostering a deeper understanding of mathematical principles among pupils/students.

In addition, there is an increasing fascination with the advancement of automated proof assistant systems, also known as interactive theorem provers (Roberts, 2023). These systems employ artificial intelligence and are projected to either equal or surpass the problem-solving proficiency of the most accomplished human mathematicians in the coming decade.

**Appendix A - Bayes Theorem**

Bayesian reasoning, named after the Reverend Thomas Bayes who contributed to the development of Bayesian probability theory, is a mathematical framework for reasoning under uncertainty. It provides a systematic way to update our beliefs probabilities as new evidence or information becomes available.

At its core, Bayesian reasoning involves using Bayes theorem to update the probability of a hypothesis based on new evidence. The theorem is described by:

$$P(A|B) = \frac{P(B) \times P(B|A)}{P(A)}$$

where:
$P(A|B)$ is the posterior probability of hypothesis $A$ given evidence $B$.
$P(B|A)$ is the likelihood of evidence $B$ given hypothesis $A$
$P(A)$ is the prior probability of hypothesis $A$ before considering evidence $B$.
$P(B)$ is the probability of evidence $B$.

The Bayesian approach allows us to update our initial beliefs (prior probabilities) about a hypothesis based on new evidence, and the result is the posterior probability, which represents our updated belief after considering the evidence.